\pdfoutput=1

\documentclass[11pt]{article}

\usepackage[]{acl}
\usepackage{url}
\usepackage{hyperref}

\usepackage{times}
\usepackage{latexsym}

\usepackage[T1]{fontenc}

\usepackage[utf8]{inputenc}

\usepackage{microtype}

\usepackage{inconsolata}

\usepackage{graphicx}
\usepackage{amsmath} 
\usepackage{bm}

\title{An Open Multilingual System for Scoring Readability of Wikipedia}

\author{Mykola Trokhymovych \\
  Pompeu Fabra University \\
  \texttt{mykola.trokhymovych@upf.edu} \\\And
  Indira Sen \\
  University of Konstanz \\
  \texttt{indira.sen@uni-konstanz.de} \\\AND
  Martin Gerlach \\
  Wikimedia Foundation \\
  \texttt{mgerlach@wikimedia.org} \\}

\begin{document}
\maketitle
\begin{abstract}
With over 60M articles, Wikipedia has become the largest platform for open and freely accessible knowledge. 
While it has more than 15B monthly visits, its content is believed to be inaccessible to many readers due to the lack of readability of its text. 
However, previous investigations of the readability of Wikipedia have been restricted to English only, and there are currently no systems supporting the automatic readability assessment of the 300+ languages in Wikipedia. 
To bridge this gap, we develop a multilingual model to score the readability of Wikipedia articles. 
To train and evaluate this model, we create a novel multilingual dataset spanning 14 languages, by matching articles from Wikipedia to simplified Wikipedia and online children encyclopedias. 
We show that our model performs well in a zero-shot scenario, yielding a ranking accuracy of more than 80\% across 14 languages and improving upon previous benchmarks. 
These results demonstrate the applicability of the model at scale for languages in which there is no ground-truth data available for model fine-tuning. Furthermore, we provide the first overview on the state of readability in Wikipedia beyond English.

\end{abstract}

\section{Introduction}
\label{sec:intro}
\let\thefootnote\relax\footnotetext{To appear in ACL’24.} 
The concept of readability aims to capture how easy it is to read a given text, usually defined as the sum of all factors that affect a reader's understanding, reading speed, and level of interest~\cite{Dale1949concept}.
In practice, the goal is often to model and quantify the readability of a text on a pre-defined scale using linguistic features, known as Automatic Readability Assessment (ARA)~\cite{Vajjala2022trends}.
Common approaches are based on readability formulas such as the Flesch-Kincaid score~\cite{Kincaid1975derivation}, with a recent shift towards more complex computational models leveraging progress on language models in NLP~\cite{Francois2015when}.
These readability scores are used to better serve readers' information needs in educational contexts for choosing appropriate reading materials to support, e.g., language learners~\cite{Xia2016text} or readers with learning disabilities~\cite{Rello2012graphical}. 
Assessing the accessibility of content in terms of readability is also of relevance more broadly, as general textual information found on the web or in the news is often linguistically too complex for large fractions of the population~\cite{Stajner2021automatic}. 
\begin{figure}[tb]
  \centering
  \includegraphics[width=0.99\linewidth]{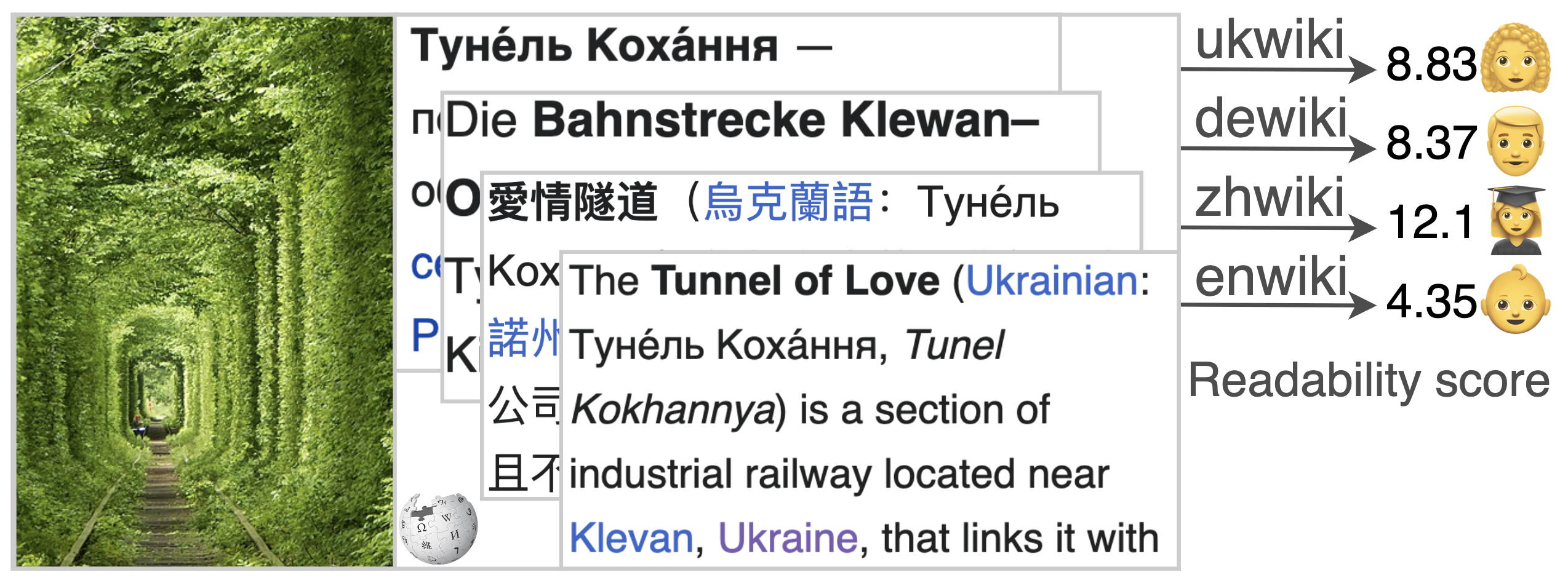}
  \caption{Sketch of the readability scoring system for Wikipedia articles. Higher scores indicate more difficult-to-read text.}
  \label{fig:idea}
\end{figure}

This use case for ARA is particularly relevant for Wikipedia, which has become the largest platform for open and freely accessible knowledge, read by millions of people worldwide with more than 60M articles across 300+ language versions~\cite{wikistats}.
Unfortunately, this knowledge is believed to remain inaccessible to many readers because the text is written at a level above their reading ability -- denoted as the readability gap~\cite{Redi2020taxonomy}.
In fact, studies on English Wikipedia have concluded that ``overall readability is poor''~\cite{Lucassen2012readability}.

However, the state of readability in Wikipedia beyond English is unknown.
Despite recent advances in ARA, there is no currently available system to systematically score articles across many languages due to several challenges (see also~\citet{Vajjala2022trends}).
There is a lack of availability of ready-to-use multilingual approaches, as existing web interfaces such as \citealp{translated} or \citealp{readable}, support only a limited number of languages. 
At the same time, there are no established readability formulas, such as the Flesch Reading Ease Formula, for most languages beyond English.
Furthermore, models (or formulas) for ARA are often designed only for individual or pairs of languages, which makes it challenging to adapt existing models because \textit{(i)} they are difficult to scale to hundreds of languages, and \textit{(ii)} resulting scores cannot be easily compared across languages~\cite{Martinc2021supervised}. 
Furthermore, there is a general lack of ground-truth data. While there are many resources for English, the datasets are often small in size and some of the most commonly-used ones are not available under an open license (such as Newsela~\cite{Xu2015problems} or WeeBit~\cite{Vajjala2012improving}), severely limiting their use in real-world applications. Beyond English, resources are scarce and scattered, such that there are no ready-to-use datasets in most languages.

In this paper, we develop a multilingual system to score the readability of articles in Wikipedia (see Figure~\ref{fig:idea}).
Specifically, we make the following contributions.
First, we compile a new multilingual dataset of pairs of encyclopedic articles with different readability levels covering 14 languages and make it publicly available under an open license.
Second, we develop a single multilingual model for readability, demonstrating the effectiveness of the zero-shot cross-lingual transfer.
Third, we apply the model to obtain the first systematic overview of the state of readability of Wikipedia articles beyond English and provide a public API endpoint of the model for use by readers, editors, and researchers.

\section{Related work}
\label{sec:related}

\paragraph{Traditional approaches} 
Research on how to measure readability dates back to more than 100 years~\cite{DuBay2007unlocking}. 
These early attempts focused mainly on developing vocabulary lists of common words and/or readability formulas, such as Flesch reading ease~\cite{Flesch1948new}, SMOG~\cite{McLaughlin1969smog}, or the Dale-Chall readability formula~\cite{Dale1948formula}, some of which are still commonly used today. In the past decades, there has been a shift towards computational models using approaches from NLP and machine learning, see the general overviews by \citet{Collins-Thompson2014computational, Francois2015when, Vajjala2022trends}.

\paragraph{Language models} 
More recently, ARA has been dominated by approaches using language models based on deep neural networks~\cite{Martinc2021supervised}. 
A wide variety of architectures have been proposed based on, among others, word embeddings ~\cite{Filighera2019automatic}, multiattentive recurrent neural networks~\cite{Azpiazu2019multiattentive}, and increasingly common transformers~\cite{Mohammadi2019text} such as BERT~\cite{Devlin2019bert}.
These models have been combined with traditional linguistic features~\cite{Deutsch2020linguistic, Imperial2021bert}.
As an alternative to the common approach of modeling ARA as a classification task, the formulation as a ranking problem has been shown to perform better in cross-corpus and cross-lingual scenarios~\cite{Lee2022neural, Miliani2022neural}.
Also, recent work utilizes prompt-based learning based on seq2seq models to solve ARA as a text-to-text generative task~\cite{lee-lee-2023-prompt}.

\paragraph{Multilingual ARA} 
While most research on ARA is focused on English, there have been many efforts in the past years for a broad range of languages such as Arabic~\cite{Nassiri2023approaches}, Cebuano~\cite{Imperial2022baseline}, Dutch~\cite{hobo2023geen},
French~\cite{Wilkens2022fabra}, German~\cite{Blaneck2022automatic}, Greek~\cite{Chatzipanagiotidis2021broad}, Spanish~\cite{Vasquez-Rodriguez2022benchmark}, or Turkish~\cite{Uluslu2023exploring}.
However, most of these studies focus only on a single language, with few exceptions attempting to model several languages jointly~\cite{Madrazo_Azpiazu2020cross, Imperial2023automatic}.
Some studies have demonstrated zero-shot cross-lingual transfer for individual pairs of languages (i.e., a model is trained only in one language and then evaluated on another) for English to French~\cite{Lee2022neural}, Spanish to Catalan~\cite{Madrazo_Azpiazu2020analysis}, or English to German~\cite{Weiss2021using}. 
Our work extends these insights beyond individual pairs, taking advantage of the more general findings that multilingual transformer models, such as mBERT~\cite{Devlin2019bert}, perform surprisingly well at zero-shot cross-lingual transfer learning for a wide range of tasks outside ARA~\cite{Pires2019how}.

\paragraph{Readability in Wikipedia} 
There have been efforts to capture readability in Wikipedia, with most studies focusing on English and using traditional readability formulas.
A comparison of Simple and English Wikipedia has shown that 
articles from Simple Wikipedia are easier to read~\cite{Yasseri2012practical}, overall readability is insufficient for its target audience (even for Simple Wikipedia)~\cite{Lucassen2012readability}, and readability in Wikipedia (both English and Simple) lags behind other encyclopedias such as Britannica~\cite{Jatowt2012wikipedia}.
\cite{Den_Besten2014keep} analyzed the temporal evolution of Simple Wikipedia, showing a gradual decline in readability.
Some studies focus specifically on Wikipedia's health-related content 
finding that most articles remain written at a level above the reading ability of average adults~\cite{Reavley2012quality,Brezar2019readability}.
Readability has also been discussed as a feature for article quality~\cite{Liu2021can, Moas2023automatic}.

\section{Data}
\label{sec:data}
We generate a new multilingual dataset of document-aligned pairs of encyclopedic articles, where each pair contains the same article in two levels of readability (easy/hard). 
The pairs are obtained by matching Wikipedia articles (hard) with the corresponding version from different simplified or children's encyclopedias~\cite{childrenswikis} (easy). 
The latter encyclopedias are purposefully designed with the goal of creating articles using simpler language (e.g. vocabulary, grammar, and sentence structure)~\cite{simplepolicy}. 
The proposed approach yields a dataset covering 14 languages summarized in Table~\ref{tab:basic_eda}.

While some of the same sources have already been used previously, e.g. Simple English Wikipedia~\cite{Zhu2010monolingual}, Klexikon~\cite{Aumiller2022klexikon}, Vikidia~\cite{Lee2022neural}, our dataset provides substantial improvements: 
i) two new data sources (\citealp{txikipedia,wikikids}));
ii) 11 new languages (Armenian, Basque, Catalan, Dutch, Greek, Italian, Occitan, Portuguese, Russian, Sicilian, Spanish);
iii) improved extraction of the plain text from articles by parsing the HTML version instead of wikitext;
iv) the dataset is publicly available under an open license in contrast to some of the most commonly used resources in ARA, such as Newsela~\cite{Xu2015problems}.

\begin{table}[tb]
\small
\begin{center}
{\tabcolsep=3pt
\begin{tabular}{l|r|r|r}
\hline
\textbf{Dataset} & \textbf{\#Pairs} & \textbf{Avg. \#Sen.} & \textbf{Avg. \#Char.} \\\hline
simplewiki-en & 112,342 & 6.2/7.9 & 84.6/130.9 \\
\hline
vikidia-en & 1,991 & 6.4/14.3 & 83.3/142.8 \\
vikidia-ca & 234 & 5.2/9.7 & 79.3/145.2 \\
vikidia-de & 260 & 6.4/11.2 & 75.8/131.0 \\
vikidia-el & 39 & 6.0/11.8 & 96.8/134.9 \\
vikidia-es & 1,915 & 5.7/7.7 & 109.0/179.4 \\
vikidia-eu & 571 & 6.5/8.7 & 114.6/129.5 \\
vikidia-fr & 12,221 & 5.7/7.3 & 106.9/152.1 \\
vikidia-hy & 485 & 14.3/11.4 & 105.3/115.1 \\
vikidia-it & 1,662 & 4.5/6.0 & 84.6/152.6 \\
vikidia-oc & 12 & 4.2/7.1 & 77.0/105.6 \\
vikidia-pt & 809 & 5.7/11.8 & 97.3/157.9 \\
vikidia-ru & 125 & 5.8/11.2 & 83.8/110.6 \\
vikidia-scn & 10 & 3.8/4.7 & 50.9/86.3 \\
klexikon-de & 2,255 & 17.7/8.9 & 73.9/136.9 \\
txikipedia-eu & 1,162 & 7.3/8.4 & 107.4/126.4 \\
wikikids-nl & 12,090 & 8.0/7.5 & 83.7/112.0 \\
\hline
\end{tabular}%
}
\caption{Dataset summary statistics (easy/hard).}
\label{tab:basic_eda}
\end{center}
\end{table}

\subsection{Dataset sources}
\label{sec:datasources}
\textbf{Simple Wikipedia} is a simplified version of English Wikipedia written in basic and learning English targeted towards children, non-native speakers, and people with learning disabilities. 
It is a commonly-used resource for large-scale text simplification datasets,
such as PWKP~\cite{Zhu2010monolingual}, SEW~\cite{Coster2011learning} WikiLarge~\cite{Zhang2017sentence}, WikiAuto~\cite{Jiang2020neural}, DWikipedia~\cite{Sun2021document}, or SWiPE~\cite{Laban2023swipe}.

\textbf{Txikipedia} is a children encyclopedia contained in the Basque Wikipedia. Similar to article talk pages, the children's version of an article is stored under a different namespace and available to readers as a separate tab at the top of the page. 

\textbf{Vikidia, Klexikon, and Wikikids} are wiki-based encyclopedias for children independent from the language versions of Wikipedia hosted by the Wikimedia Foundation (WMF). Vikidia exists in 11 different languages. \citet{Azpiazu2019multiattentive} considered six of the languages in their experiments, but the article-aligned data is not publicly available. \citet{Lee2022neural} compiled data for English and French.
Klexikon is available in German and was utilized by \citet{Aumiller2022klexikon} to create text simplification datasets, and Wikikids is available in Dutch.

\subsection{Pre-processing}
\label{sec:preprocessing}
We match articles from Wikipedia with the corresponding article in the simplified/children encyclopedia either via the Wikidata item id or their page titles.
We extract the text of each article directly from their parsed HTML version instead of using the original wikitext~\cite{wikitext}, the markup language in which Wikipedia is edited. 
Previous studies show that this approach provides a more accurate representation of the content of Wikipedia articles as seen by its readers~\cite{Mitrevski2020wikihist}. 
For example, using wikitext as a source often results in missing important information~\cite{wikitexthugginghace}.
In order to limit systematic differences in length, we only consider the text from the first (lead) section.
We only keep pairs of articles in which both versions of the text have three or more sentences.
For more details about data processing, see Appendix~\ref{sec:appendix_preprocessing}.

\section{Model}
\label{sec:model}

\subsection{Design requirements}
In this section, we describe the requirements for the system to score the readability of Wikipedia articles across languages, as they influence the architecture of the underlying model.

First, our aim is to score the readability of articles in Wikipedia on a continuous scale. 
Typically, ARA is modeled as a classification problem in NLP research with the goal to predict the labels of the ground-truth data corresponding to a few readability levels (e.g., five in Newsela or three in OneStopEnglish). 
In our case, the intended use-case is not to predict the label corresponding to the article's source (Wikipedia or simplified/children encyclopedia) but to score articles on a more fine-grained scale.

Second, we aim to develop a single multilingual model. 
This will not only allow us to compare scores across different languages, but also reduce infrastructure costs related to scaling and maintaining the model for many languages.

Third, the model should require no or only little language-specific fine-tuning (i.e. zero-shot or few-shot scenario) as there is little to no annotated ground-truth data on readability available for almost all of the 300+ languages in Wikipedia.

\subsection{Model architecture}
\label{sec:model_arc}
\begin{figure}[tb]
  \centering
  \includegraphics[width=0.99\linewidth]{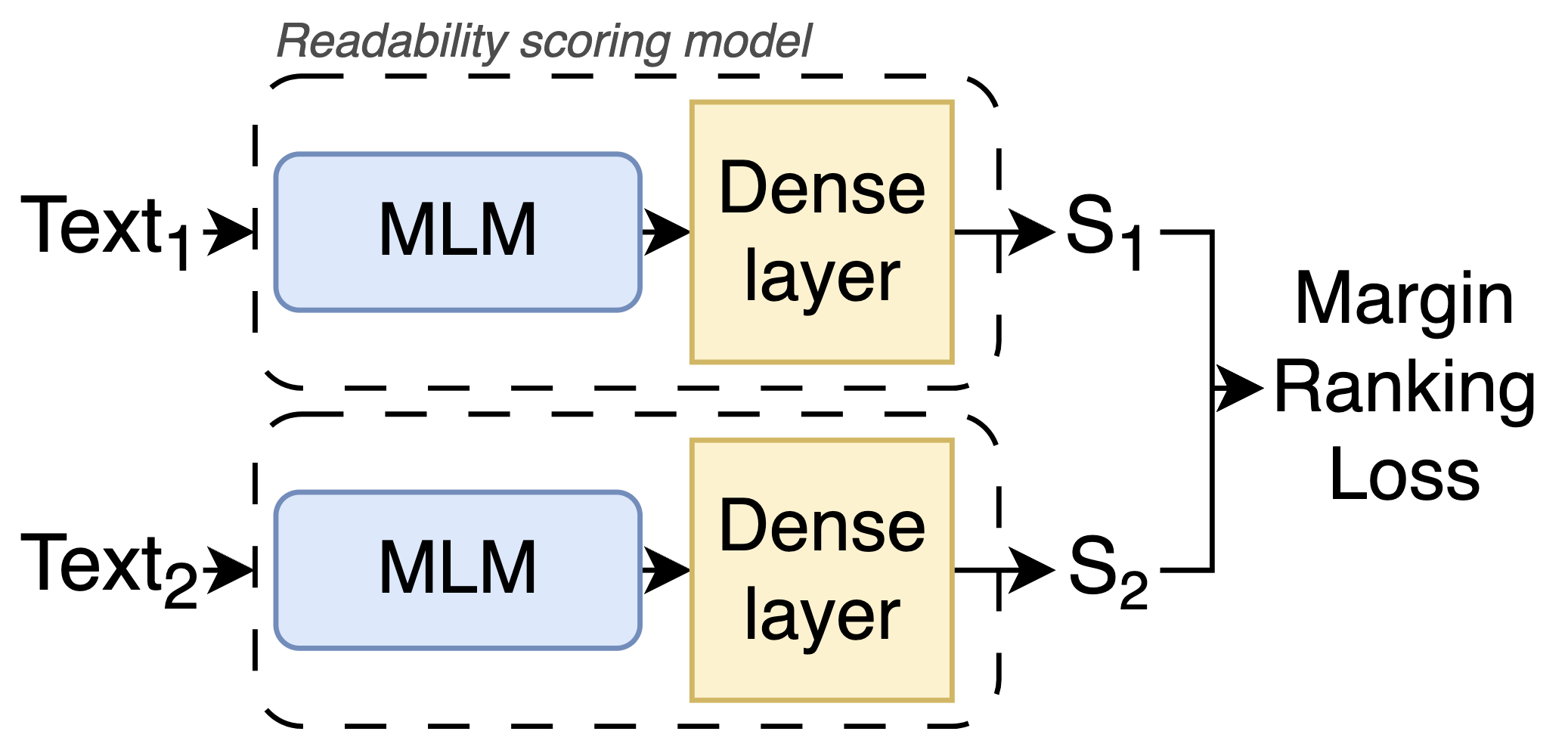}
  \caption{Sketch of the model architecture consisting of two joint readability scoring models trained using a Margin Ranking Loss. $S_1$ and $S_2$ refer to the predicted scores of $Text_1$ and $Text_2$, respectively.
  }
  \label{fig:model_training}
\end{figure}

We train a ranking-based model that can perform text scoring of individual texts during inference (see Figure~\ref{fig:model_training} for a sketch of the training procedure).  
For this, we adapt the Neural Pairwise Ranking Model (NPRM) introduced by \citet{Lee2022neural} to overcome its main limitation with respect to scoring the readability of articles, i.e. that the model requires the input of at least two texts and only provides a relative ranking as output.
We first build a readability scoring model composed of a multilingual Masked Language Model (MLM)~\cite{Devlin2019bert} and a Dense layer. 
The MLM takes the text as input and encodes it into a numerical representation. 
The Dense Layer then performs a linear transformation into a single real number, which serves as a readability score.
We then use a Siamese architecture~\cite{siamese} composed of two joint readability scoring models that share their weights.
We apply a Margin Ranking Loss (MRL), also known as Pairwise Hinge Loss, that is commonly used for ranking models, such as Ranking SVM~\cite{herbrich1999support}),  given by: 
$$\text{MRL}(S_1, S_2, y) = \max(0, -y(S_1 - S_2) + \text{m}),$$
where $m$ is a hyperparameter. 
The MRL is a function of a pair of the same text annotated with readability levels, where $S_1$ and $S_2$ are the predicted scores of the texts and $y=-1$ or $1$ depending on whether the first or second text should have a lower or higher score based on the annotation. 
During inference, we pass each individual text to the readability scoring model to obtain the score that is used for readability assessment. 

\subsection{Fine-tuning strategy}
We fine-tune the model for ARA using the dataset consisting of pairs of encyclopedic articles available in two readability levels (Sec.~\ref{sec:data}).
More precisely, we use only the \textit{simplewiki-en} dataset for fine-tuning, splitting the data randomly into a training ($80\%$) and testing ($20\%$) dataset.
All other datasets are only used for testing.

This is motivated by the fact that the main goal is to apply the model without language-specific fine-tuning.
The \textit{simplewiki-en} dataset is by far the largest available annotated dataset for ARA, providing a large volume and wide spectrum of training examples.
Using a multilingual MLM as the backbone of our model, which has been pre-trained in an unsupervised setting in more than 100 languages, we expect zero-shot cross-lingual transfer learning~\cite{Pires2019how}.

\subsection{Technical implementation}
We utilize the transformers package~\cite{wolf-etal-2020-transformers} to fine-tune the model. 
We implement the ranking-based readability model in two different flavors.

The text-based model (TRank) takes as input the full text of each article at once. 
For this, we use the \textit{xlm-roberta-longformer-base}~\cite{Sagen1545786} as a backbone, as it allows us to process long inputs of up to 4096 tokens. 

As an alternative, we also implement a sentence-based version (SRank), where the text is split into sentences that are passed to the model sequentially, and as a result, leading to much smaller input lengths. 
For this, we use the \textit{bert-base-multilingual-cased}~\cite{Devlin2019bert} as a backbone. 
The main motivation for adding SRank experiments alongside the TRank model is that the model is smaller (i.e. requiring fewer computational resources during inference) and that it can, in principle, process articles of any length without truncation (addressing a key limitation of the TRank model).

Details about the hyperparameters and computational resources can be found in Appendix ~\ref{sec:appendixB}.

\section{Experimental evaluation}
\label{sec:evaluation}

\subsection{Task setup}

\paragraph{Ranking task}
We evaluate the model in a pair-wise ranking task following the approach in~\cite{Lee2022neural}. 
That is, for each pair of articles, we assume that the model's readability score should be lower for the easy text (from the simplified/children encyclopedia) than for the hard text (from Wikipedia). 
This approach has several advantages over a binary classification task aiming to predict the readability label: 
\textit{(i)} we take advantage of document-level alignment of the same text in different readability levels instead of predicting labels of individual documents; 
\textit{(ii)} pair-wise ranking directly evaluates the model's readability scores by checking whether the easy version receives a lower score.
As an evaluation metric, we use ranking accuracy (RA), that is the percentage of pairs that are ranked correctly. 
We use bootstrapping to compute confidence intervals (see Appendix~\ref{sec:appendixC}).

\paragraph{Baselines}
In order to evaluate our model's performance, we consider a set of strong baselines that are representative of the most common approaches.

\textit{NS} constitutes a naive baseline that calculates the number of sentences in each text.

\textit{FRE} calculates the Flesch reading ease of the text.
Using \citealp{textstat}, we obtain the language-specific reading ease formula when available and English-specific otherwise.

\textit{LFR and LFC} constitute a ranker and classifier, respectively, which are based on linguistic features. We use the LFTK tool~\cite{Lee2023lftk} to extract all language-agnostic features (via the parameter \textit{language=``general''}). Using these features, we then train a ranker and classifier model using CatBoost~\cite{catboost} with default parameters for 5K iterations with a learning rate of $0.01$.

\textit{LMC} uses a multilingual MLM for a classification-based approach (instead of ranking-based). We fine-tune the \textit{bert-base-multilingual-cased} model~\citet{Devlin2019bert}, splitting the texts into sentences and applying mean pooling to get aggregated readability scores.

\subsection{Results}

\paragraph{Multilingual benchmark}
We evaluate the performance of our model and the baselines on our new multilingual benchmark dataset (Table~\ref{tab:ranking_accuracy}). 
Overall, our model substantially outperforms all baselines, yielding an RA above $0.8$ across all datasets, and generally with TRank being slightly better than the SRank model.
In the following, we inspect these results in more detail.

First, we observe that our model (TRank) yields an almost perfect RA of $0.976$ on the test set of \textit{simplewiki-en}.
Taking into account that we fine-tuned the model on the corresponding training set, this performance might not be surprising. In fact, many of the baselines yield RA of around 0.9 or higher.

Second, we consider performance on a different dataset not used for training but still in the same language as the training data (\textit{vikidia-en}).
We observe that most models yield RA above 0.979, demonstrating that the models generalize well beyond the specific training data.

Third, we consider performance in languages that were not used for fine-tuning the model (zero-shot scenario).
Our model (TRank) yields an RA higher than $0.8$ for all datasets and higher than $0.9$ for 10 out of 15 non-English datasets.
Notably, the SRank model performs substantially worse in some of the languages (e.g., Basque). 
In comparison, the performance of the baseline model varies substantially across languages. 
The naive \textit{NS} baseline yields generally poor results across most languages. 
The  \textit{FRE} baseline performs well in English, but RA in other languages is substantially lower than for our models.
The RA for the \textit{LFC, LFR, LMC} baselines is similar to our models in some cases (e.g. \textit{vikidia-it}) but much lower for others, most notably languages with non-Latin scripts (\textit{vikidia-el}, \textit{vikidia-ru}, \textit{vikidia-hy}).

As the TRank model outperforms SRank across almost all datasets and languages, we will consider only the TRank model for all experiments in the remainder of the paper. 

\begin{table*}[t]
\begin{center}
\small
{\tabcolsep=3pt
\begin{tabular}{l|rl|rl|rl|rl|rl|rl|rl}
\hline
\textbf{Dataset} & \textbf{NS} & \textbf{±CI} & \textbf{FRE} & \textbf{±CI} & \textbf{LFC} & \textbf{±CI} & \textbf{LFR} & \textbf{±CI} & \textbf{LMC} & \textbf{±CI} & \textbf{TRank} & \textbf{±CI} & \textbf{SRank} & \textbf{±CI} \\
\hline
simplewiki-en & 0.543 & 0.007 & 0.868 & 0.005 & 0.937 & 0.003 & 0.945 & 0.003 & 0.965 & 0.002 & \textbf{0.976} & 0.002 & 0.972 & 0.002 \\\hline
vikidia-en & 0.814 & 0.017 & 0.935 & 0.011 & 0.979 & 0.006 & 0.981 & 0.006 & 0.979 & 0.006 & \textbf{0.991} & 0.004 & 0.985 & 0.005 \\
vikidia-ca & 0.782 & 0.054 & 0.906 & 0.038 & 0.94 & 0.031 & 0.932 & 0.033 & 0.936 & 0.032 & \textbf{0.962} & 0.025 & 0.936 & 0.032 \\
vikidia-de & 0.735 & 0.054 & 0.815 & 0.048 & 0.888 & 0.039 & 0.869 & 0.042 & 0.908 & 0.036 & \textbf{0.938} & 0.03 & 0.919 & 0.034 \\
vikidia-el & 0.718 & 0.144 & 0.718 & 0.144 & 0.744 & 0.14 & 0.795 & 0.129 & 0.897 & 0.096 & \textbf{0.923} & 0.086 & 0.897 & 0.097 \\
vikidia-es & 0.573 & 0.023 & 0.842 & 0.017 & 0.883 & 0.015 & 0.892 & 0.014 & 0.879 & 0.015 & \textbf{0.911} & 0.013 & 0.909 & 0.013 \\
vikidia-eu & 0.541 & 0.042 & 0.673 & 0.04 & 0.639 & 0.04 & 0.623 & 0.041 & 0.63 & 0.04 & \textbf{0.818} & 0.032 & 0.736 & 0.037 \\
vikidia-fr & 0.553 & 0.009 & 0.84 & 0.007 & 0.82 & 0.007 & 0.845 & 0.006 & 0.849 & 0.007 & \textbf{0.923} & 0.005 & 0.918 & 0.005 \\
vikidia-hy & 0.394 & 0.045 & 0.594 & 0.045 & 0.534 & 0.045 & 0.598 & 0.044 & 0.637 & 0.044 & \textbf{0.802} & 0.036 & 0.761 & 0.039 \\
vikidia-it & 0.569 & 0.024 & 0.83 & 0.019 & 0.919 & 0.013 & 0.94 & 0.012 & 0.925 & 0.013 & \textbf{0.958} & 0.01 & 0.952 & 0.01 \\
vikidia-oc & 0.667 & 0.273 & 0.667 & 0.271 & 0.75 & 0.25 & 0.667 & 0.27 & 0.917 & 0.159 & \textbf{1.0} & 0.0 & 0.917 & 0.161 \\
vikidia-pt & 0.761 & 0.03 & 0.869 & 0.024 & 0.938 & 0.017 & 0.934 & 0.017 & 0.921 & 0.019 & \textbf{0.960} & 0.014 & 0.938 & 0.017 \\
vikidia-ru & 0.728 & 0.08 & 0.608 & 0.087 & 0.736 & 0.078 & 0.776 & 0.074 & 0.736 & 0.079 & \textbf{0.880} & 0.058 & 0.760 & 0.077 \\
vikidia-scn & 0.4 & 0.314 & 0.6 & 0.309 & 0.6 & 0.308 & 0.8 & 0.254 & 0.6 & 0.31 & 0.9 & 0.191 & \textbf{1.0} & 0.0 \\ \hline
klexikon-de & 0.114 & 0.013 & 0.984 & 0.005 & 0.999 & 0.002 & 0.995 & 0.003 & 0.991 & 0.004 & \textbf{0.999} & 0.002 & 0.996 & 0.003 \\
txikipedia-eu & 0.512 & 0.029 & 0.707 & 0.027 & 0.689 & 0.027 & 0.698 & 0.027 & 0.67 & 0.027 & \textbf{0.81} & 0.023 & 0.762 & 0.025 \\
wikikids-nl & 0.427 & 0.009 & 0.795 & 0.007 & 0.831 & 0.007 & 0.834 & 0.007 & 0.85 & 0.007 & 0.897 & 0.006 & \textbf{0.907} & 0.005 \\
\hline
\end{tabular}%
}
\caption{Ranking accuracy on different test datasets (zero-shot scenario for all datasets except simplewiki-en). Confidence intervals (CI) denote two standard deviations from bootstrapping (Appendix~\ref{sec:appendixC}).}
\label{tab:ranking_accuracy}
\end{center}
\end{table*}

\paragraph{Reference datasets}
In order to directly compare our results with previous work, we also evaluate our model on three open reference datasets considered in the experiments by~\citet{Lee2022neural}, who introduced the $NPRM$ as one of the most recent SOTA approaches in multilingual ARA:  Vikidia-En, Vikidia-FR, and OneStopEnglish (OSE)~\cite{Vajjala2018onestopenglish}.
For the latter dataset with 3 readability levels, RA is the fraction of triples where predicted scores for all three versions are ranked correctly.

In Table~\ref{tab:benchmarks}, we show the performance of our model on the corresponding test datasets in a zero-shot scenario, i.e. without any additional training or fine-tuning.
The results show that TRank substantially outperforms not only the baselines but also the $NPRM$, especially in French.
Surprisingly, many of the baselines yield high RA (e.g. simplistic features using the number of sentences in VikidiaEn and VikidiaFr). 
This further highlights the usefulness of our proposed multilingual benchmark, as it seems to constitute a more challenging dataset for ARA tasks.

\begin{table*}[tb]
\small
\begin{center}
{
\tabcolsep=3pt
\begin{tabular}{l|rl|rl|rl|rl|rl|rl|rl}
\hline
\textbf{Dataset} & \textbf{NS} & \textbf{±CI} & \textbf{FRE} & \textbf{±CI} & \textbf{LFC} & \textbf{±CI} & \textbf{LFR} & \textbf{±CI} & \textbf{LMC} & \textbf{±CI} & \textbf{TRank} & \textbf{±CI}  & \boldsymbol{$NPRM$} & \textbf{±CI}\\
\hline
VikidiaEn & 0.966 & 0.005 & 0.948 & 0.006 & 0.888 & 0.008 & 0.946 & 0.006 & 0.965 & 0.005 & \textbf{0.984} & 0.003 &0.975 & 0.004\\
VikidiaFr & 0.952 & 0.005 & 0.899 & 0.008 & 0.878 & 0.008 & 0.888 & 0.008 & 0.75 & 0.011 & \textbf{0.978} & 0.004 &0.811 & 0.010\\
OSE & 0.794 & 0.059 & 0.915 & 0.04 & 0.889 & 0.046 & 0.873 & 0.048 & 0.942 & 0.034 & \textbf{0.974} & 0.023 &0.878& 0.048\\
\hline
\end{tabular}%
}
\end{center}
\caption{Ranking accuracy on previous reference datasets (zero-shot scenario). Results of the $NPRM$ model taken from~\cite{Lee2022neural}. Confidence intervals (CI) denote two standard deviations from bootstrapping (Appendix~\ref{sec:appendixC}). }
\label{tab:benchmarks}
\end{table*}

\subsection{Interpreting the model's readability scores}
\label{sec:interpretation}

Our model yields a readability score on a continuous scale, which can take positive and negative values and higher scores indicate that the text is more difficult to read.

We find that the readability scores are strongly and statistically significantly correlated ($p$-value < $10^{-12}$) with existing readability formulas adapted for different languages. 
Specifically, we calculate the Spearman rank correlation between the model's readability scores and the language-specific Flesch reading ease (FRE)~\cite{textstat} for articles in the corresponding languages: 
$-0.63$ (simplewiki-en) and $-0.72$ (vikidia-en), $-0.67$ (vikidia-de) and $-0.81$ (klexikon-de), $-0.67$ (vikidia-es), $-0.65$ (vikidia-fr), $-0.76$ (vikidia-fr), $-0.62$ (wikikids-nl), and $-0.43$ (vikidia-ru).\footnote{A negative correlation is expected as higher FRE scores indicate easier-to-read texts. }
These results demonstrate that the readability scores of our multilingual model correspond to existing and well-founded notions of readability across languages. 

\begin{figure}[tb]
  \centering
  \includegraphics[width=0.99\linewidth]{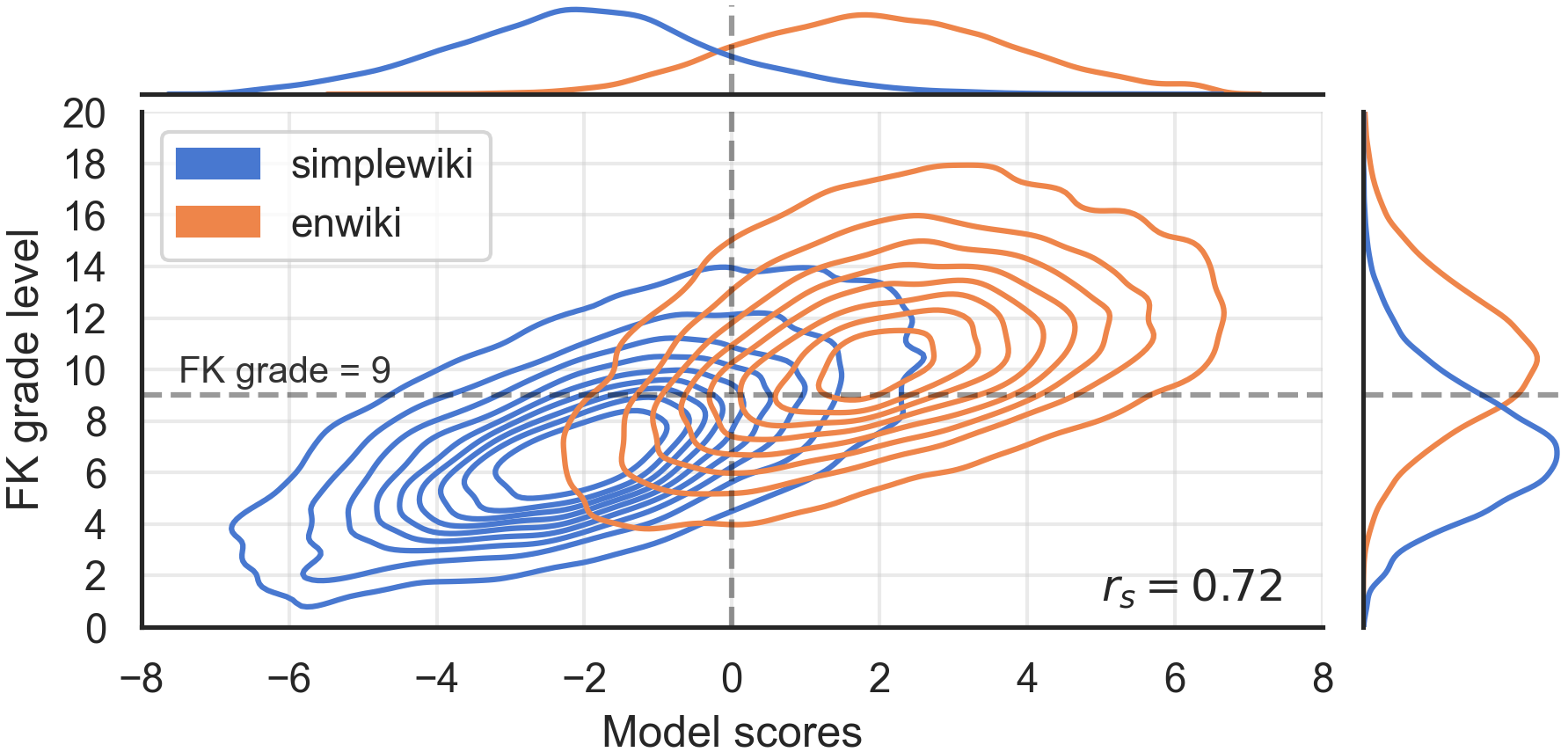}
  \caption{Distribution of model scores vs. FKGL for articles from the test set of simplewiki-en stratified by readability level: simplewiki (easy) and enwiki (hard).}
  \label{fig:joinplot}
\end{figure}

Focusing on English, we associate the model's readability score with the Flesch-Kincaid grade level (FKGL)~\cite{Kincaid1975derivation}. The main advantage of FKGL is its interpretability in terms of U.S. grade level or, loosely speaking, the number of years of education required to understand a text. Higher grade levels indicate more difficult text. Using articles from simplewiki-en, we find that scores are significantly and strongly correlated ($\rho=0.72$, $p$-value < $10^{-12}$).
In Figure~\ref{fig:joinplot}, we show that, as expected, a model score of $\approx0$ separates texts from simplewiki (easy) and enwiki (hard).
We find that this separation corresponds to an FKGL $\approx9$.
This allows us to roughly map the model's reference point to a corresponding grade level and interpret the readability of articles that are above or below.
 
\section{Application}

\subsection{State of readability in Wikipedia}
We use the TRank model to get an overview on the state of readability in Wikipedia beyond English.
For this analysis, we consider overall 24 different Wikipedias covering all languages from our dataset (Sec.\ref{sec:data}) in addition to a set of 10 languages considered in a prior study covering different language families and taking into account the number and distribution of speakers worldwide~\cite{Lemmerich2019why}.
For each language, we select a random subsample of 10K articles and extract the text following the same methodology as in Sec.~\ref{sec:preprocessing}.

\begin{figure}[tb]
  \centering
  \includegraphics[width=0.98\linewidth]{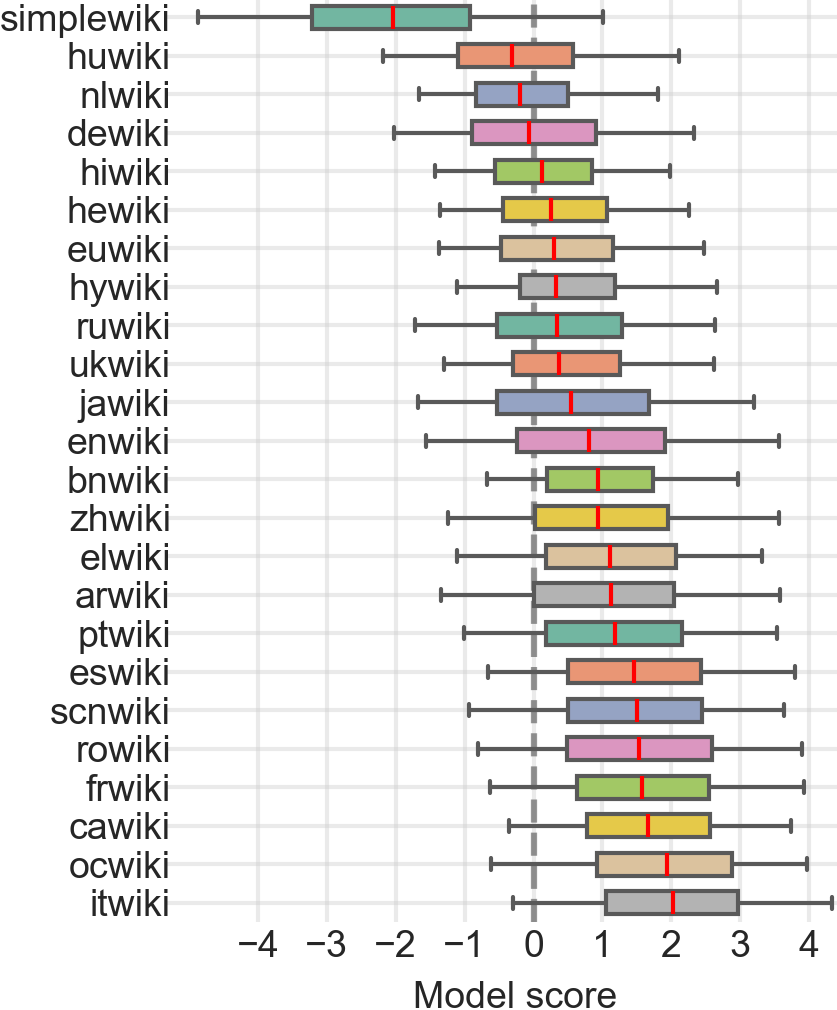}
  \caption{Distribution of readability scores (from the TRank model) across different language editions of Wikipedia. Boxplots show median (red line) and 25- and 75-percentiles with whiskers ranging from 2.5- to 97.5-percentile.}
  \label{fig:boxplot}
\end{figure}

In Figure~\ref{fig:boxplot}, we show the distribution of readability scores across articles in each language.
For English Wikipedia, we observe a median score around $1$, with the majority of articles above a score of 0 (roughly corresponding to a Flesch-Kincaid grade level of 9 (Sec.~\ref{sec:interpretation}).
For most languages, the distribution of scores is similar or shifted towards much higher values, such as Italian, with a median of $2.02$. 
Some languages show slightly lower scores, such as Hungarian, with a median of $-0.32$.
Only Simple Wikipedia, as expected, shows substantially lower readability scores with a median of $-2.04$ and the $75$-percentile below $0$.
Overall, this demonstrates that previous findings about the poor overall readability of English Wikipedia~\cite{Lucassen2012readability} can be generalized to most other language editions of Wikipedia.

\subsection{Productionization as a ready-to-use tool}
In order to facilitate the use of our model, we provide a public API endpoint to directly access readability scores for articles in Wikipedia from the model.
The deployed model is an end-to-end system, including the articles' text collection using the MediaWiki API, processing, and scoring.\footnote{\url{https://api.wikimedia.org/wiki/Lift_Wing_API/Reference/Get_readability_prediction}}

We measure efficiency on inference by selecting a random sample of 1K articles and passing them sequentially to the readability model. 
We observe the median response time of $0.5$ seconds, and the $75\%/95\%$ percentiles are $0.83/2.05$ seconds when limiting resources to 1 thread CPU. 
The same measurements are $0.02/0.03/0.05$ when GPU is enabled. It should be mentioned that inference time is influenced by the length of text, as long articles require more processing by the language model. 

\section{Discussion}
\label{sec:discussion}
\subsection{Summary of findings}
We created a new dataset for multilingual ARA with pairs of aligned articles in two readability levels from Wikipedia and a corresponding simplified/children encyclopedia. 
The advantage of our dataset is that it i) contains new sources (such as Txikipedia) and languages (such as Basque), ii) provides cleaner plain text from processing HTML sources, and iii) is publicly available under an open license.\footnote{\url{https://zenodo.org/records/11371932}} 

We developed a new multilingual model for ARA, adapting a ranking-based architecture to score individual texts across languages. 
We demonstrate that the model performs well in the zero-shot scenario across all languages with a ranking accuracy $>0.8$, substantially outperforming all baselines, including traditional readability formulas.
This suggests that the model can be applied at scale to languages in which we do not have ground-truth data for additional fine-tuning.
We provide additional insights into the interpretability of this model's readability scores by showing that they correlate with hand-crafted readability formulas available for individual languages.
In order to ensure reproducibility, we make the code for training and evaluating the model available in a public repository.\footnote{\url{https://gitlab.wikimedia.org/repos/research/readability-experiments}}

We apply our model to get the first state of readability in Wikipedia beyond English. 
We reproduce previous findings in English Wikipedia: most of the content is written at a reading level above the average (American) adult~\cite{Lucassen2012readability, Brezar2019readability}.
We find that the readability of most languages in Wikipedia is at a similar (or even more difficult) level than English Wikipedia.
In order to facilitate the use of our model, we provide a publicly available API endpoint to the trained model for researchers and contributors as a ready-to-use tool for ARA in Wikipedia.

\subsection{Implications and broader impact}
\paragraph{Children encyclopedia communities} 
Our work shows that the simplified and children encyclopedias provide an invaluable resource for multilingual research on readability. 
More generally, it highlights that these encyclopedias, which are targeted towards specific audiences, play an important role in the open online knowledge ecosystem. 
However, very little is known about these projects. 
In order to better understand its content (e.g. quality, reliability, suitability for different readability levels), more research is needed about its audience (who is using it) and contributors (who is creating it) and their motivations. 

\paragraph{Improving the state of multilingual ARA}
Our results improve the state of ARA by directly addressing some of its main limitations according to one of the most recent reviews on the topic~\cite{Vajjala2022trends}.
First, we address the lack of publicly available multilingual corpora by providing a new, high-quality, multilingual dataset under an open license.
Second, our model addresses the lack of availability of ready-to-use tools: We not only provide the code in a public repository together with documentation in the form of a model card, but also make available a public API endpoint for users.
Third, our results address the lack of well-defined SOTA: We provide reproducible new benchmarks for 14 languages (datasets and training/evaluation code).

\paragraph{The extent of the readability gap in Wikipedia}
Our tool provides a systematic approach to quantify readability as a knowledge gap in Wikipedia~\cite{Redi2020taxonomy}. 
We start from the observation in English Wikipedia that the readability scores of the majority of articles exceed a reading level corresponding to a Flesch-Kincaid grade level of 9 (Sec.~\ref{sec:interpretation}).
This means that much of its content is not accessible to the larger population in terms of readability when taking into account that the average reading ability of adult Americans is estimated at grade 7-8~\cite{Mcinnes2011readability} (matching recommendations for readability levels of public resources such as for patient education material by the U.S. National Institutes of Health ~\cite{Brezar2019readability}).
Our results show that these insights can be generalized to most other language versions, as the distribution of articles' readability scores is similar or shifted towards higher difficulty.

More generally, this expands previous research on motivations of readers~\cite{Singer2017why, Lemmerich2019why}, which has shown that information needs vary substantially with demographics such as gender~\cite{Johnson2021global}.
Measuring the readability of articles describes the suitability of content for readers with different educational and/or literacy backgrounds. 
In this way, it is possible to identify misalignment between supply (readability of existing content) and demand (education levels of the potential reader population) in Wikipedia, similar to previous studies focusing on information quality~\cite{WarnckeWang2015misalignment}.

\paragraph{Text simplification}
Our model provides a starting point for systematically approaching the task of text simplification in Wikipedia in order to make content more accessible to different audiences.
The use of ARA for the automatic evaluation of text simplification approaches~\cite{AlvaManchego2020data} can now be applied across languages.
Also, ARA can identify those articles that are the most difficult to read (and thus, the most in need of simplification).
Surfacing those to contributors would enable them to make data-informed editorial decisions taking into account readability~\cite{wishlist}. 

\section*{Limitations}
We tried two variants of MLMs (SRank and TRank), and found them to have similar performances. Similar larger models with more parameters, such as mLongT5~\cite{Uthus2023mlongt5}, could yield even better performance. 
However, the necessary infrastructure (especially in terms of GPUs) required for training and inference makes it challenging to provide the model as a ready-to-use tool.

Multilingual models based on transformer architectures support many languages (e.g., multilingual BERT was trained on 104 languages). 
However, among those supported, the performance on low-resource languages is still considered unsatisfactory~\cite{Wu2020all}.
More severely, the majority of the more than 300 languages in Wikipedia is still not explicitly represented in the training data of these models. 
Thus, if unaddressed, the use of such models could lead to a language gap constituting a substantial barrier towards knowledge equity~\cite{Redi2020taxonomy}. 

We evaluated our multilingual model on only 14 languages for which we were able to compile a ground truth dataset of encyclopedic articles available at different readability levels. 
It should be noted that the models were trained only on English texts, so the scores for unseen languages constitute approximations.
Additional validation in an applied scenario~\cite{Vajjala2022trends}, beyond showing statistically significant correlations with commonly-used language-specific readability formulas, would be desirable for future research using, e.g., comprehension tests such as Cloze tests~\cite{Redmiles2019comparing}.

Our models and experiments focus only on document-level readability assessment, evaluating the overall readability of entire articles. This approach differs from other forms of ARA that target finer-grained levels, such as sentence-level or phrase-level readability. By concentrating on document-level assessments, we aim to provide a general readability score for Wikipedia articles, though this may overlook variations in readability within smaller sections of text.

\section*{Ethics statement}
We develop multilingual datasets and models for measuring the readability of Wikipedia articles to better understand the state of readability on Wikipedia, across its many language editions. 
By pinpointing articles with low readability scores, we support editors and researchers in identifying and addressing these gaps in order to make knowledge on Wikipedia more accessible. 

\textbf{Dataset Quality }
We contribute a novel and openly available dataset of encyclopedic articles covering different readability levels. It is the largest of its kind and covers 11 more languages compared to past work. By making it available under an open license, we provide a valuable resource for NLP researchers, especially those working on ARA.

Wikipedia articles have been used to create a wide variety of NLP datasets, especially those used to train large language models~\cite{gao2020pile}. 
Our dataset consists of encyclopedic articles from Wikipedia and other online encyclopedias. 
While the dataset contains some metadata about the articles (e.g., link to Wikidata), it does not contain any details about the author(s) of the articles. 
Therefore, the dataset does not divulge any private information about the articles' authors or readers. 
Furthermore, some of the online encyclopedias, especially those hosted by WMF, have robust community-driven moderation that ensures that the content is reliable~\cite{yasseri2021can}. 
We also take further steps in processing and filtering (cf. Section~\ref{sec:preprocessing}), to improve data quality, a crucial issue in multilingual NLP research~\cite{kreutzer2022quality}. 

In terms of language variety, while we do not have fine-grained information about the editors writing the articles in this dataset, previous research on the demographics of the editors of English Wikipedia have revealed that they are primarily men from North America and Europe~\cite{communityinsights}.
However, Wikipedia is \textit{read} by a more diverse group of people~\cite{readerdemographics}. So, we expect this dataset and the models built using it to be useful for this more diverse global audience. 

\textbf{Intended Use of Models }
We also develop a model for scoring the readability of Wikipedia articles. Since there are few resources for assessing the readability of non-English text, even less so for low-resourced languages, one of the main strengths of our model is its promising zero-shot cross-lingual transfer capabilities. This model is hosted and publicly deployed so that it can be easily used in an off-the-shelf manner, without investing effort in training the model from scratch. We intend for this model to be used not only by researchers interested in investigating the state of readability in Wikipedia articles across languages but also by other stakeholders such as readers and editors. By assessing the current status of readability in Wikipedia, editors can flag articles needing further simplification. 

We do not intend for this dataset and model to be used to increase or reinforce biases, e.g., to discriminate against people whose writing is automatically scored with lower readability scores, or to profile or censor people based on their writing. 
This model was tested for encyclopedia-like text and might not generalize to other forms of writing, such as academic assignments. 
In future research, we hope to use the readability scores from our multilingual model in tandem with other metrics of knowledge accessibility, such as visual content and citations, to meet the needs of Wikipedia readers.

\section*{Acknowledgements}
The work of Mykola Trokhymovych is funded by MCIN/AEI /10.13039/501100011033 under the Maria de Maeztu Units of Excellence Programme (CEX2021-001195-M).

\bibliography{custom}
\newpage
\appendix

\section{Additional data characteristics}
\label{sec:appendixA}
\begin{figure}[b]
  \centering
  \includegraphics[width=0.99\linewidth]{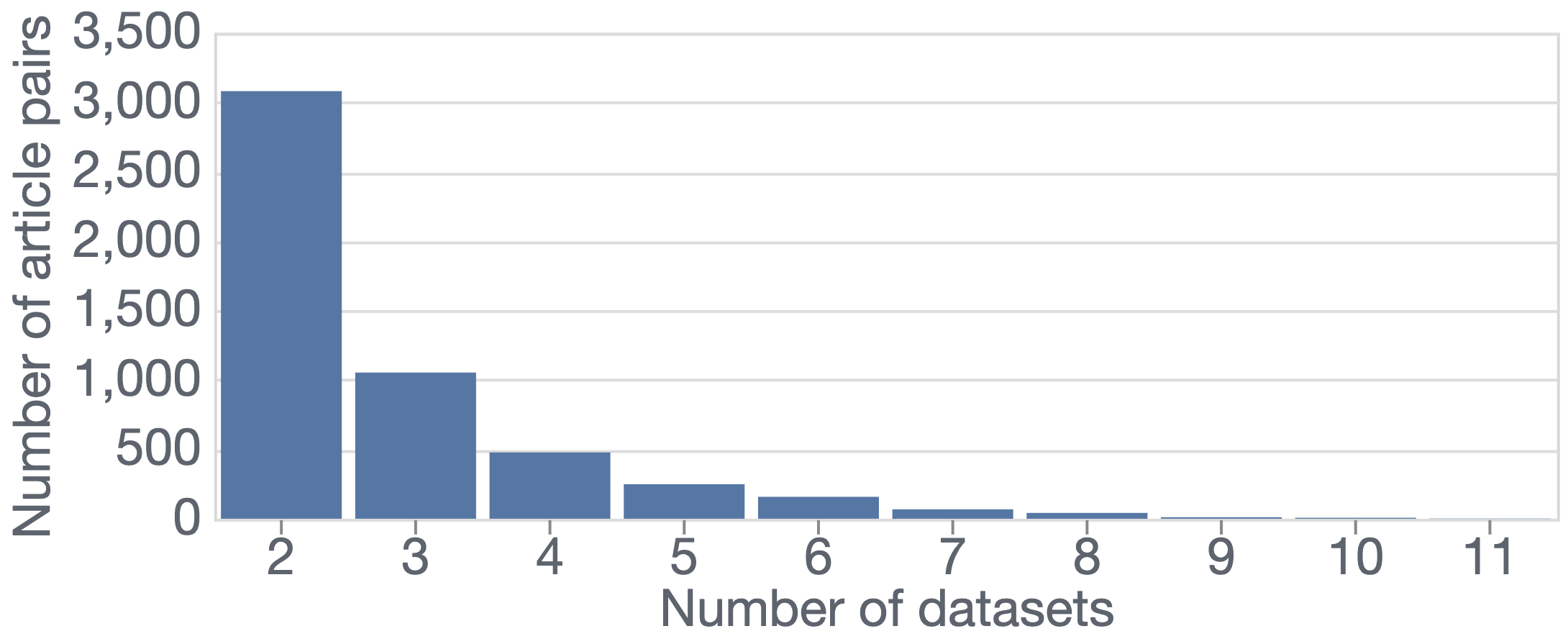}
  \caption{Number of articles that occur in two or more different datasets (single occurrence is skipped).}
  \label{fig:coocurance}
\end{figure}

The new dataset presented consists of pairs of hard and simple text of one article for multiple languages. However, we also mention that there are articles that are present in multiple languages. For example, we have more than 3K article pairs that are presented in two languages (Figure~\ref{fig:coocurance}).

\section{Data preprocessing}
\label{sec:appendix_preprocessing}
We start by matching articles from Wikipedia with the corresponding article in the simplified/children encyclopedia.
For simplewiki, we match all articles from English Wikipedia and Simple English Wikipedia via their Wikidata item ids. We remove all disambiguation and list pages as they often display only itemized lists without continuous text.
For Txikipedia, we match page titles of all articles in namespace 0 (Main namespace) and 104 (Txikipedia), respectively; for example, the article ``Klima'' will have ``Txikipedia:Klima'' as the title for the children's version. 
For Vikidia, Klexikon, and Wikikids, we match the page title in Wikipedia with the corresponding article in the children encyclopedia. Due to different naming conventions and the fact that the two sources are not completely aligned, articles might have different titles (such as ``Baby'' and ``Infant'' in English Wikipedia and Vikidia, respectively). To address this, we also consider all redirects of an article as alternative titles and match pairs if we find one and only one match between all titles of an article.

We get the HTML-version of each article from the HTML dumps~\cite{htmldumps}
or, alternatively for sources in which they are not available, the Wikimedia APIs (\citealp{restapi},\citealp{actionapi}).

We then parse the HTML of the article to extract the plain text. 
We first split the article into sections, only keeping the first section of each article (lead section) to limit extreme differences in length between the two versions. 
We then only consider text within \verb|<p>|-tags to avoid text from, e.g., infoboxes, image captions, etc.
Within each \verb|<p>|-tag, we extract the plain text from each element, removing any formatting (e.g. from links) and removing text that is from references or in sub/super-script.
This leads to much cleaner plain text than if one would parse the wikitext, mainly due to the wide usage of templates~\cite{Mitrevski2020wikihist}. 
While there exist packages for expanding the content of templates in wikitext, such as WikiExtractor~\cite{Wikiextractor2015}, they require the full dump files, which are not publicly available for the children encyclopedias that are not WMF-hosted. 

We only keep pairs of articles in which each version consists of three or more sentences to limit the number of low-quality or stub articles. 

Note that, for each pair of articles, we keep the Wikidata item id associated with the corresponding Wikipedia article. 
This allows us to align pairs of articles across different datasets and, thus, also languages (see Figure~\ref{fig:coocurance} in the Appendix). 
For example, the Wikidata item id \verb|Q433| corresponds to the pair of articles with titles ``Phyiscs'' in Vikidia (English) and ``Physik'' in Klexikon (German).

\section{Additional modeling details}
\label{sec:appendixB}

\subsection{Model hyperparameters}
For TRank, we utilize pretrained \textit{
xlm-roberta-longformer-base} ($\sim$281M parameters) model by~\citet{Sagen1545786} as a backbone. 
This model is designed to process long text, accommodating up to 4096 tokens. However, we limit it to 1500 to fit into our available resource constraints, which allows us to tokenize 99.95\% of our dataset without truncation. We concatenate all sentences to construct an article text and use it as model input. This input is then passed through a sequence of MLM and the Dense layer to generate the readability score.

We train a model for three epochs with an initial learning rate of $10^{-5}$ and weight decay equal to $10^{-7}$. Also, we use the CosineAnnealingLR scheduler with hyperparameters $min\_lr= 10^{-7}$ and $T\_max = 256$. Due to memory constraints, the batch size during training is equal to one. Also, we fix the hyperparameter $m=0.5$, which refers to the margin in the loss function. Also, we use the 1\% sample from training data as the validation set. We track the loss and select the checkpoint created after the epoch when the model shows the best performance on validation data. As a result, we get a single model with the lowest loss. It takes $\sim$80 GPU hours for training and inference of the TRank model on our dataset, and doing additional experiments presented in the paper. 

For SRank, we use \textit{bert-base-multilingual-cased} ($\sim$178M parameters) by \cite{Devlin2019bert} as a backbone. We use the same hyperparameters for model training as for the TRank model, except for batch size, which is instead set to 16 for SRank. During the inference stage, each sentence from the article is individually passed through the model, and the scores are aggregated using mean pooling.  We use the same training strategy as for the TRank model, but we prepare a custom sentence-based training dataset. To construct this dataset, we use Levenshtein Distance 
to select similar sentence pairs from the aligned articles available in different readability levels. It takes $\sim$30 GPU hours for training and inference of the SRank model on our dataset, and doing additional experiments presented in the paper. 

The choice of values for the hyperparameters was motivated by previous approaches using similar models that have been shown to perform well in practice\cite{debarshichanda}.

\subsection{Computational resources}
The model choice is motivated by the available computational resources (1x AMD Radeon Pro WX 9100 16GB GPU). 
In total, $\sim$160 GPU hours are needed to reproduce the experiments presented in this paper. 

\section{Confidence intervals}
\label{sec:appendixC}

We estimate confidence intervals of the ranking accuracy metric via bootstrapping~\cite{efron1994introduction}.
Specifically, we resample each test set 10K times by drawing randomly with replacement $N$ samples from the test set, where $N$ is the size of the test set. 
For example, if we have 1K observations in the testing data, we randomly choose 10K samples of size 1K (with replacement) from the given data. 
We then calculate the standard deviation of the ranking accuracy over the 10K bootstrap-samples.
We report two standard deviations as the confidence interval, denoted as CI in Tables~\ref{tab:ranking_accuracy} and~\ref{tab:benchmarks}.

We calculate the CI for the $NPRM$ model taken from~\citet{Lee2022neural} without reproducing model inference. 
We use the scores for each sample of benchmark datasets published by the authors, which allows us to bootstrap their results without reproducing model inference.

\end{document}